\documentclass[lettersize,journal]{IEEEtran}
\usepackage{amsmath,amsfonts}
\usepackage{algorithmic}
\usepackage{array}
\usepackage{textcomp}
\usepackage{stfloats}
\usepackage{url}
\usepackage{verbatim}
\usepackage{graphicx}
\usepackage{rotating}
\usepackage{cite}
\usepackage[ruled,vlined]{algorithm2e}
\usepackage{adjustbox}
\usepackage{multirow}
\usepackage{tablefootnote}
\usepackage{xcolor}
\usepackage[font=small,labelfont=bf]{caption}
\usepackage{listings}
\usepackage{booktabs}
\usepackage{subcaption}
\usepackage[pagebackref,breaklinks,colorlinks]{hyperref}
\usepackage[capitalize]{cleveref}
\usepackage{microtype}      
\usepackage[T1]{fontenc}    
\usepackage{nicefrac}       
\usepackage{bm}
\usepackage[utf8]{inputenc} 
\usepackage{url}            
\usepackage{amsfonts}       
\usepackage{textgreek}
\usepackage{array}
\usepackage{enumitem}
\usepackage{diagbox}
\usepackage{xfrac}
\usepackage{capt-of}
\usepackage{bm}
\usepackage{tikz}
\usepackage{comment}
\usepackage{amsmath,amssymb} 
\usepackage{color}
\usepackage[accsupp]{axessibility}  

\def\BibTeX{{\rm B\kern-.05em{\sc i\kern-.025em b}\kern-.08em
    T\kern-.1667em\lower.7ex\hbox{E}\kern-.125emX}}
\usepackage{balance}

\crefname{section}{Sec.}{Secs.}
\Crefname{section}{Section}{Sections}
\Crefname{table}{Table}{Tables}
\crefname{table}{Tab.}{Tabs.}

\newcommand{\ournas}{\text{PredNAS}}
\newcommand{\ourmodel}{PredNAS-}
\newcommand{\std}[1]{\color{black}{\tiny{$\pm$#1}}}
\newcommand{\tablestyle}[2]{\setlength{\tabcolsep}{#1}\renewcommand{\arraystretch}{#2}\centering\footnotesize}
\definecolor{codegreen}{rgb}{0,0.6,0}
\definecolor{codegray}{rgb}{0.5,0.5,0.5}
\definecolor{codepurple}{rgb}{0.58,0,0.82}
\definecolor{backcolour}{rgb}{0.95,0.95,0.92}

\lstdefinestyle{mystyle}{
    backgroundcolor=\color{backcolour},   
    commentstyle=\color{codegreen},
    keywordstyle=\color{magenta},
    numberstyle=\tiny\color{codegray},
    stringstyle=\color{codepurple},
    basicstyle=\ttfamily\footnotesize,
    breakatwhitespace=false,         
    breaklines=true,                 
    captionpos=b,                    
    keepspaces=true,                 
    numbers=left,                    
    numbersep=5pt,                  
    showspaces=false,                
    showstringspaces=false,
    showtabs=false,                  
    tabsize=2
}

\begin{document}
\def\x{{\bm x}}
\def\a{{\bm a}}
\def\MF{{\mathcal F}}
\def\MC{{\mathcal C}}
\def\MP{{\mathcal P}}
\def\x{{\bm x}}
\def\a{{\bm a}}
\def\MF{{\mathcal F}}
\def\MC{{\mathcal C}}
\def\MP{{\mathcal P}}
\def\MBV{{\mathbf{V}}}
\def\MBJ{{\mathbf{J}}}
\def\MBW{{\mathbf{W}}}

\title{$\ournas$: A Universal and Sample Efficient Neural Architecture Search Framework} 
\author{Liuchun Yuan, 
Zehao Huang,
Naiyan Wang
\thanks{L. Yuan, Z. Huang, N. Wang are work for Tusimple Inc. Email: ylc0003@gmail.com, zehaohuang18@gmail.com, winsty@gmail.com}}     

\maketitle

\begin{abstract}
In this paper, we present a general and effective framework for Neural Architecture Search (NAS), named $\ournas$. The motivation is that given a differentiable performance estimation function, we can directly optimize the architecture towards higher performance by simple gradient ascent. Specifically, we adopt a neural predictor as the performance predictor. Surprisingly, $\ournas$ can achieve state-of-the-art performances on NAS benchmarks with only a few training samples (less than 100). To validate the universality of our method, we also apply our method on large-scale tasks and compare our method with RegNet on ImageNet and YOLOX on MSCOCO. The results demonstrate that our $\ournas$ can explore novel architectures with competitive performances under specific computational complexity constraints. 

\end{abstract}
\section{Introduction}
\label{sec:intro}

Recent years have witnessed the extraordinary success of deep neural networks (DNNs) in various fields such as computer vision and neural language processing. From AlexNet \cite{krizhevsky2012imagenet} to ResNet\cite{he2016deep} and MobileNet\cite{howard2017mobilenets}, DNNs become more and more compact and efficient, which shows the importance of ``network engineering''. 
To reduce the enormous effort involved in hand-crafted network design, recent work \cite{zoph2016neural, liu2018darts, guo2020single, ofa} focus on Neural Architecture Search (NAS), which is a technique for discovering better network architecture under constrained resources automatically.

The pioneering work \cite{zoph2016neural} showed NAS could design better network architecture than hand-crafted deep models. However, the search cost is immense and unaffordable. Therefore, recent researches on NAS paid much attention to improve the search efficiency. Several work \cite{zoph2018learning, chen2019renas, hu2020angle, Radosavovic_2020_CVPR, liu2018progressive} focused on efficient sampling and progressive search. Nevertheless, the number of architectures needed to train and evaluate is still large. 
Another stream of work such as one-shot methods \cite{bender2018understanding,guo2020single} and gradient-based algorithms \cite{liu2018darts} adopted the strategy of SuperNet and weight-sharing to save the cost of model training. Among them, SPOS \cite{guo2020single} and DARTS \cite{liu2018darts} are two representatives. They both train a supernet which contains all models in the search space as subnets. Then the performance of each subnet is used as the reference of corresponding model. Differently, SPOS only uses the SuperNet for fast evaluation, while DARTS utilizes gradients to update architecture parameters along with the SuperNet training. However, these methods usually suffer from unfair sampling~\cite{chu2021fairnas} and unstable training~\cite{chen2020stabilizing,chu2020fair,chen2019progressive} issues due to the weight-sharing strategy. Moreover, the search space may be limited by GPU memory since it needs to store the whole computational graph of SuperNet in GPU. 
The last category of work relates to our method most, which adopts a neural network, generally named predictor, for fast performance evaluation \cite{wen2020neural,dai2020fbnetv3,zhang2021acenas}. The main cost of predictor based methods is the collection of training samples for the predictor, namely the architecture-accuracy pairs. To handle this problem, recent researchers proposed several techniques such as ranking-based loss functions \cite{zhang2021acenas}, data augmentation \cite{liu2021homogeneous} and predictor pre-training \cite{dai2020fbnetv3, zhang2021acenas} to improve sample efficiency. Even though, hundreds of samples are still needed, which hinders the use of predictor based methods in large scale network search.

In this paper, we focus on neural predictor based methods and propose a general and efficient framework, named $\ournas$. Our method enjoys the both advantages of predictor and gradient based NAS methods, and significantly reduce the number of training samples (a.k.a the times of training networks). Moreover, our method is general enough to accommodate different types of architecture and connection search. This property makes our method a flexible and universal tool for AutoML. 
Specifically, we randomly sample a small number of models and train a predictor for performance estimation. Then the predictor can be seen as a differentiable function which models the relationship between network architecture encoding and its corresponding performance on a specific task. Then given an initial architecture encoding $\a$, we can optimize a new architecture $\a'$ starting from $\a$ by the predictor guided gradient ascent. With this simple search strategy, our method can find new architectures with comparable performance with the state-of-the-art methods on several NAS benchmarks \cite{dong2021nats} using less than 100 training samples. To validate the universality of our method, we also conduct experiments on large scale tasks, such as image classification on ImageNet dataset \cite{deng2009imagenet} and object detection on MSCOCO \cite{lin2014microsoft}. On ImageNet, we adopt our method on the largest unconstrained AnyNet search space proposed by \cite{Radosavovic_2020_CVPR}, and show that our $\ournas$ can find a series of models with comparable performances to RegNet without shrinking the search space by human heuristics. In object detection task, we validate our method on a recent work YOLOX \cite{ge2021yolox}. The results demonstrate that our method can explore better models under different FLOPs constraints. 

To summarize, the contributions of our work are in the following two folds:
\begin{itemize}
	\item We propose a general and efficient framework for predictor based neural architecture search. With a simple gradient based search strategy, our method can adapt to various NAS problems with less than 100 training samples. 
	\item With the same framework, we conduct comprehensive experiments in various applications and search spaces to show the universality of our method. The results show that our $\ournas$ could consistently achieve comparable or even better performance than existing methods.
\end{itemize}

\section{Related Works}
\label{sec:related_works}
\subsection{Sampling based Neural Architecture Search}
To the best of our knowledge, \cite{zoph2016neural} is the pioneering work of NAS. In \cite{zoph2016neural}, a recurrent network, also named controller, is used to propose new network architectures for evaluation. The controller is trained by reinforcement learning to maximize the performance of sampled networks on specific task. Though \cite{zoph2016neural} showed the proposed method could design state-of-the-art models, the search cost was prohibitive. Several following works \cite{zoph2018learning, liu2018progressive, hu2020angle, Radosavovic_2020_CVPR} tried to mitigate this problem by reducing the search space. Different from searching the whole network on target dataset, \cite{zoph2018learning} proposed to search stackable cells on proxy tasks. PNAS \cite{liu2018progressive} adopted a progressive search approach to reduce the search cost. \cite{Radosavovic_2020_CVPR} focused on the design of search space and they discovered several principles to progressively simply the search space. Similarly, ABS \cite{hu2020angle} presented an angle-based metric to drop candidates during search. 

\subsection{Weight-sharing based Neural Architecture Search}
To alleviate the high cost of evaluation of each sampled architecture in NAS, some other works \cite{bender2018understanding, pham2018efficient, liu2018darts, guo2020single, cai2018proxylessnas, ofa, yu2020bignas} utilized the weight-sharing mechanism. The key idea is to allow the samples to share weights to reduce the training cost of each architecture. The first work in this spirit is ENAS \cite{pham2018efficient}. Then one-shot methods \cite{bender2018understanding, guo2020single, liu2018darts} proposed to train a SuperNet capable of enumerating the child models in the search space for fast model evaluation. SPOS \cite{guo2020single} adopted an uniform sampling strategy for the training of SuperNet and used evolutionary algorithm for architecture search. DARTS \cite{liu2018darts} and DSO-NAS \cite{zhang2020you} formulated NAS as an optimization problem and optimized the architecture parameters by gradient descent. However, SuperNet based methods may suffer from the problem of unreliable ranking \cite{chu2021fairnas} and performance collapse \cite{zela2019understanding, chen2020stabilizing, chen2019progressive}. FairNAS \cite{chu2021fairnas} proposed a strict fairness sampling and training strategy to improve the ranking correlation of SPOS. \cite{zela2019understanding} proposed to early stop the search process to handle the instability problem of DARTS. SmoothDARTS \cite{chen2020stabilizing} found the rounding step of deriving the discrete architecture from continuous optimization in DARTS could introduce large performance drop. They proposed a perturbation-based regularization to smooth the loss landscape of DARTS.

\subsection{Predictor based Neural Architecture Search}
In the context of hyperparameter optimization, early work adopted probabilistic models \cite{domhan2015speeding} or Bayesian neural networks \cite{klein2016learning} to estimate the performance of neural network. The subsequent work tried to predict the accuracy of neural networks by neural networks. Peephole \cite{deng2017peephole} used an LSTM \cite{hochreiter1997long} network to integrate the information of different layers following the network topology and adopted a Multiple Layer Perceptron (MLP) to predict the accuracy of the input network at a specific training epoch. \cite{wen2020neural} proposed to use Graph Convolutional Networks (GCNs) \cite{kipf2016semi} to represent the connections of network architecture. The network operations are represented as one-hot codes and the topology of the neural network is formulated by a adjacency matrix. FBNetV3 \cite{dai2020fbnetv3} adopted a simple MLP to search both architectures and training recipes jointly. Additionally, they proposed a pre-training approach to improve the sample efficiency of predictor. Recently, instead of improving the regression accuracy of predictor, ReNAS \cite{xu2021renas} and AceNAS \cite{zhang2021acenas} adopted rank based loss for reliable predictions. The closest work to our PredNAS is NAO\cite{luo2018neural}, which adopted gradient to do architecture optimization. However, NAO utilized encoder-decoder framework to do transformation between architecture and network embedding. Hundreds of training samples and an extra structure reconstruction loss are needed for the training. In our framework, the design of network encoding and projection function allows us to do architecture embedding transformation without encoder and decoder. The training samples used in our work (30) in much less than NAO (600), and we show our PredNAS can directly search on large-scale tasks such as ImageNet without transferring the searched architecture from CIFAR to ImageNet.

\subsection{NAS Applications on Different Tasks}
Designing the structure of backbone network on image classification task and then transfer the ImageNet pretrained model to downstream tasks is a de facto paradigm in computation vision community. However, recent works show that this paradigm may be sub-optimal due to the gap between classification task and downstream applications. DetNAS \cite{chen2019detnas} adopted the technique of one-shot SuperNet for object detection backbone search. Auto-DeepLab \cite{liu2019auto} constructed a hierarchical architecture search space for semantic segmentation task and adopted gradient-based method for search.  SpineNet \cite{du2020spinenet} proposed a scale-permuted model and showed the learned backbone could achieve better performance than regular scale-decreased models on both object detection and image classification tasks. Other works adopted NAS for component search, such as the architecture of feature pyramid \cite{ghiasi2019fpn} and prediction head \cite{wang2020fcos} in object detection. In this paper, we adopted our $\ournas$ on a very efficient object detection framework, YOLOX \cite{ge2021yolox}, and show we can explore better architectures based on a large search space proposed by us.

\section{Method}
\label{sec:method}
In this section, we will first introduce the motivation of our method, and then describe the framework of $\ournas$ step by step, including the formulation of search space and the details of predictor training and architecture search.

\subsection{Motivation}
Given the architecture search space $\mathbf{\Omega}$ and target resource budget $c$,
NAS can be formulated as the following optimization problem:
\begin{equation}\label{equ:optimization}
\a^* = \operatorname*{argmax}_{\a\in \mathbf{\Omega}} \ \MF(\a) , \operatorname{s.t.} \  \MC(\a) <= c,
\end{equation}
where $\MF(\a)$ is the performance indicator of architecture $\a$ on a specific dataset and $\MC(\a)$ is the corresponding resource cost such as FLOPs, latency or parameters. For simplicity, we denote $\a$ as the encoding of its corresponding architecture. Following \cite{tan2019mnasnet}, we are more interested in finding multiple Pareto-optimal \cite{deb2014multi} solutions instead of a single architecture with highest accuracy. We construct a formulation of weighted sum of objectives to approximate Pareto optimal solutions:
\begin{equation}\label{equ:pareto_optimization}
\a^* = \operatorname*{argmax}_{\a\in \mathbf{\Omega}} \ \MF(\a) - \alpha (\MC(\a) - c),
\end{equation}
where $\alpha$ is a tunable weight. 
Then the key problem becomes: \emph{1) How could we approximate $\MF(\cdot)$ and $\MC(\cdot)$ with limited training data? 2) How could we optimize the above problem efficiently?} Our answer is surprisingly simple: \emph{neural network with gradient descent}\footnote{\scriptsize{Though the formulation of $\MC(\cdot)$ with FLOPs (or parameters) guided constraints can be derived directly, it is difficult to formulate latency or power guided computational complexity in a parametric form. For a general  formulation, we adopt a predictor to approximate $\MC(\cdot)$.}}. In this paper, we adopt two neural networks, the main predictor and the auxiliary predictor to approximate $\MF(\cdot)$ and $\MC(\cdot)$ respectively. Then the gradient of $\a$ respect to $\MF(\a)$ and $\MC(\a)$ can be computed by back-propagation easily for the optimization of $\a$. 

\subsection{Search Space and Network Encoding}
\label{sec:method_ss}

\begin{figure}[t]
\begin{minipage}{0.5\textwidth}
    \centering
    \includegraphics[width=.9\linewidth]{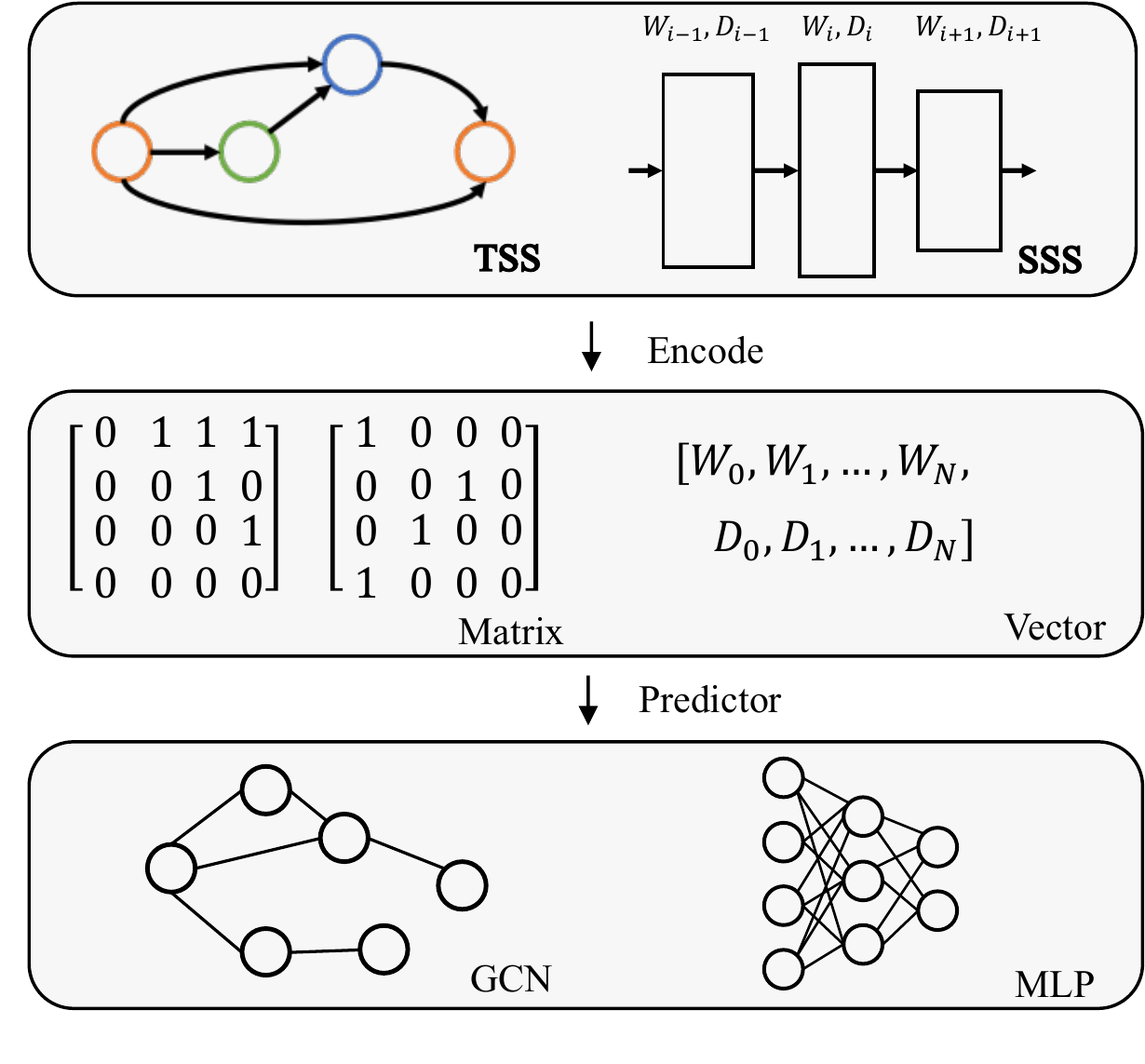}

\end{minipage}
\begin{minipage}{0.5\textwidth}
	\centering
	\scriptsize
	\setlength\extrarowheight{6pt}
	\begin{tabular}{|l|l|c|l|l|}
		\hline
		Method &  Type  & Pred. & Num.  & Task  \\
		\hline
		\hline
		AnyNet\cite{Radosavovic_2020_CVPR} & size  & MLP &  $\approx 10^{18}$ & cls.  \\
		YOLOX\cite{ge2021yolox}\ & size  &   MLP   &  $\approx 10^{100}$ & det. \\ 
		Bench-201\cite{dong2019bench} & topology &  GCN & $\approx 10^{5}$ & cls.\\
		\hline
		\hline
		SPOS\cite{guo2020single}  &   size   & MLP  & $\approx 10^{12}$ & cls. \\ 
		DARTs\cite{liu2018darts}  & topology & GCN  & $\approx 10^{9}$  & cls. \\ 
		NAS-FPN\cite{ghiasi2019fpn} & topology &  GCN & $\approx 10^{16}$ & det.\\ 
		\hline
	\end{tabular}

\end{minipage}
\captionsetup{font=scriptsize}
\caption{\textbf{Top}: The Process of Different search space. \textit{First row}, the macro skeleton of each architecture. \textit{Second row}, the architecture encoding methods. \textit{Third row}, the network structures of predictors. \textbf{Bottom}: The search space and predictors used in different tasks. The top is methods compared in \ournas. The bottom is other methods. }
\label{fig:typeenc}

\end{figure}

Defining search space is the first step in NAS. Generally, the search space in NAS can be divided into two categories\cite{dong2021nats} as shown in Fig.~\ref{fig:typeenc}(left): (1) \textbf{Topology Search Space (TSS)} adopted in DARTS\cite{liu2018darts} and NAS-Bench-201\cite{dong2019bench}, concerning the connection topology and the associated operations on the connections; (2) \textbf{Size Search Space (SSS)} focuses on the width, depth or other parameters with the same topology \cite{guo2020single, Radosavovic_2020_CVPR, ofa}. Following \cite{dai2020fbnetv3} and \cite{wen2020neural}, we adopt different network encoding methods to represent the architectures sampled from these two search space. For TSS, an adjacency matrix is used to represent the topology of architecture, and the operations on each connections are encoded as an operation matrix which consists of one-hot vectors. As for SSS, we simply concatenate the values of network depths, channel widths and other architecture parameters into a vector. 

\begin{figure*}[t!]
	\centering
	\includegraphics[width=.9\linewidth]{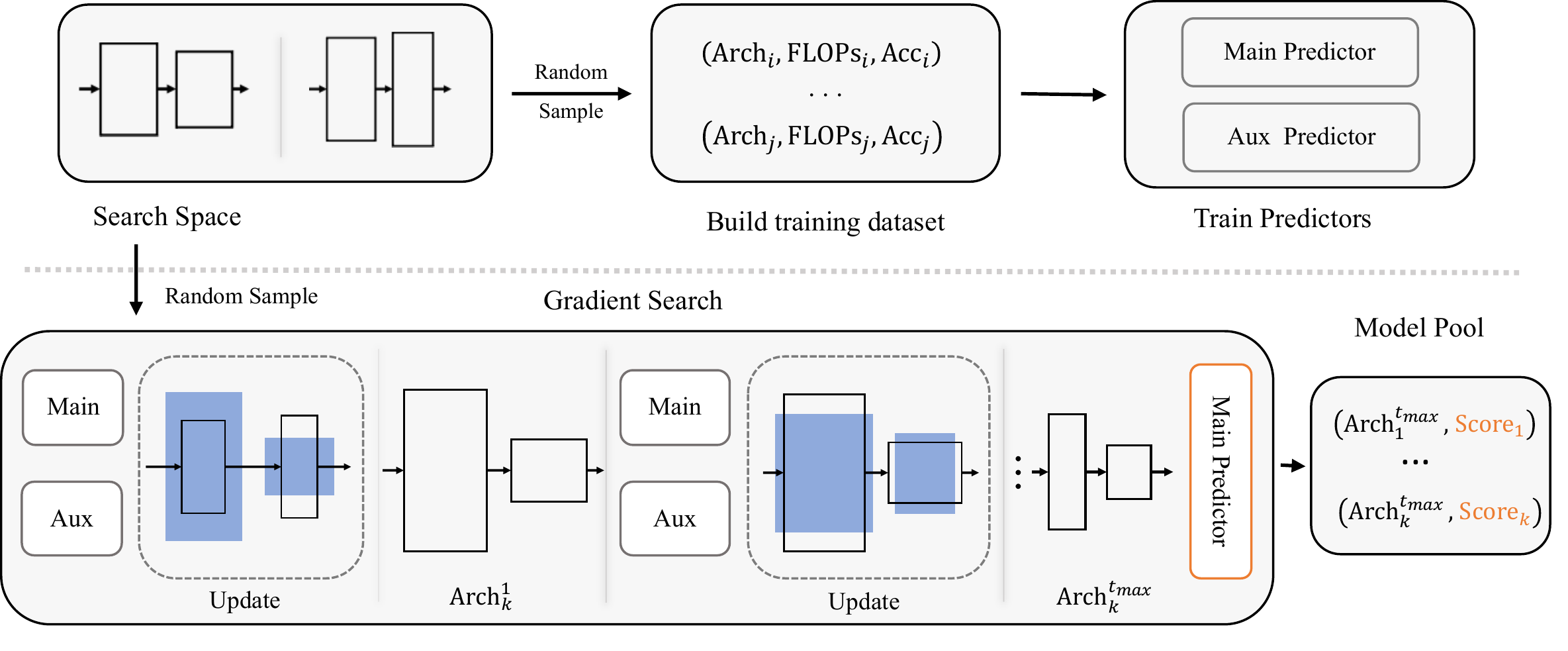}
	
	\caption{The framework of \ournas. \textbf{Top}: the training process; \textbf{Bottom}: gradient-based search with predictors. During training, we collect some training samples to train the main and auxiliary predictors. As for search, we adopt the gradient of predictors to progressively update the widths, depths and other parameters of the random sampled architectures. The blue color rectangle means the architecture of one layer in the network after gradient update. Figure best viewed in color.}
	\label{fig:frame}
	\vspace{-1em}
\end{figure*}
\subsection{Training Predictors}
\label{sec:method_train}
We train two predictors to approximate $\MF(\cdot)$ and $\MC(\cdot)$: (1) \textbf{main predictor} $\MP_m$, estimating the performance of given architectures, e.g.\ accuracy or mean average precision (mAP); (2) \textbf{auxiliary predictor} $\MP_{aux}$, predicting the resource cost of given architectures. These two predictors share the same network structure. For encoding of network structure, it highly depends on the search space of each dataset, so we leave it in the experiment section. For TSS, we adopt Graph Neural Network (GCN) \cite{kipf2016semi} to handle graph-structured training samples following \cite{wen2020neural}. As for SSS, we use a simple MLP.

The top of Fig.~\ref{fig:frame} shows the training process. For simplicity, we consider FLOPs constraints and size search space for illustration in the following paper. We first randomly sample $N$ ($N<=50$) architectures from search space, and then collect the training samples as a dataset of $(\a, acc., \text{FLOPs})$ triplets. The performance of architecture can be obtained via training the architecture on the target task or a proxy task to reduce the overhead. Following \cite{dai2020fbnetv3}, we adopt the Huber loss to train both the main and auxiliary predictors.

\subsection{Gradient-based Search with Predictors}
\label{sec:method_search}
In the search stage, most of the predictor based methods \cite{wen2020neural,dai2020fbnetv3,zhang2021acenas} adopt predictors only for architecture evaluation. They usually randomly sample a large number (about $10^6$) of architectures, and then take the networks with top-K performances to retrain. In this paper, we argue that \emph{these methods essentially utilize naive random shooting method for optimization, and ignore the valuable differentiable property of the predictor}. Actually, we can update the sampled architecture according to the gradient of predictors. The bottom of Fig.~\ref{fig:frame} shows the search procedure of our $\ournas$. Given an initial architecture $\a$, the main predictor $\MP_m$ and the auxiliary predictor $\MP_{aux}$, we update $\a$ iteratively by projected gradient ascent:
\begin{equation}\label{equ:gradientupdate}
	\a^{(t+1)} = \mathbf{P}_\mathbf{\Omega}(\a^{(t)} + \eta * (\frac{\partial{\MP_m(\a^{(t)})}}{\partial{\a^{(t)}}} -\alpha \frac{\partial{\MP_{aux}(\a^{(t)})}}{\partial{\a^{(t)}}})),
\end{equation}
where $\eta$ is the learning rate, $t$ is the number of iterations, $\alpha$ is a trade-off between performance and resource constraints and $\mathbf{P}_\mathbf{\Omega}$ is a projection function which projects the updated architecture embedding back onto search space $\mathbf{\Omega}$. 
	We formulate $\mathbf{P}_\mathbf{\Omega}$ as the following optimization problem:
	\begin{equation} 
	\mathbf{P}_\mathbf{\Omega}(\bm{a}^{(t)})=\operatorname*{argmin}_{\bm{a}\in \mathbf{\Omega}}{\frac{1}{2}}||\bm{a} - \bm{a}^{(t)}||.
	\label{equ:projection}
	\end{equation}
	In size search space, the output of this projection function can be obtained by rounding and clipping. As for topology search space, we select the operation with maximum probability as the chosen operation.
 Updating $\a^{(t)}$ with gradient $\frac{\partial{\MP_m(\a^{(t)})}}{\partial{\a^{(t)}}}$ increases the performance of $\a^{(t)}$ while $\frac{\partial{\MP_{aux}(\a^{(t)})}}{\partial{\a^{(t)}}}$ reduces the computational complexity. With a suitable $\alpha$, the gradient-based search strategy will optimize the architecture $\a$ towards to higher performance while reducing the resource cost. Given a number of initial architectures, we can explore new networks with higher performances based on the proposed gradient based search method efficiently. To obtain architectures under a hard resource constraints, we adopt a grid search strategy to tune the value of $\alpha$ and all the models under the target constraints during update will be added into the model pool.

Alg. \ref{alg:search_algorithm} describes the whole process of search. After training predictors, we randomly sample $T_{max}$ architectures from search space. Then we update each architecture following Eqn.(\ref{equ:gradientupdate}) until reaching max iteration $t_{max}$. We collect all models within constraints to a model pool and sort them by the predictor scores. Finally, we will retrain top-K of them on a specific task and select the best model as the final result.

\begin{algorithm}[ht]
	\footnotesize
	\SetAlgoLined
	\textbf{Input:} Size search space $\mathbf{\Omega} \in \mathbb{R}^n$, target FLOPs interval ${[f-\delta:f+\delta]}$, \\
	the loss weight $\alpha$, the main and auxiliary predictors $\MP_m$, $\MP_{aux}$ and learning rate $\eta$\\
	ModelPool=[];\\
	\For{$T=1,2,...,T_{max}$}{
		Random sample initial architecture $\a^{(1)} \in \Omega$\;
		$\a^{(1)}$.requires\_grad = True\; 
		\For{$t=1,2,...,t_{max}$}{
			$\text{loss} = - \MP_m(\a^{(t)}) + \alpha * \MP_{aux}(\a^{(t)})$\;
			$\text{loss.backward()}$\;
			$\a^{(t+1)} = \a^{(t)} - \eta * \a^{(t)}.\text{grad}$\;
			$\a^{(t+1)} \leftarrow \mathbf{P}_\mathbf{\Omega}(\a^{(t+1)})$\;  
			calculate FLOPs of $\a^{(t+1)}$ and get $f_{\a^{(t+1)}}$\;
			\If { $f_{\a^{(t+1)}} \in [f-\delta:f+\delta]$}{
				score = $\MP_m(\a^{(t+1)})$\; 
				ModelPool.append$\{(\a^{(t+1)}, \text{score})\}$\;
			}
		}
	}
	Sort ModelPool according to score in  descending order\;
	\Return ModelPool[:K]
	\caption{FLOPs guided gradient search algorithm in \ournas}
	\label{alg:search_algorithm}
\end{algorithm}

\section{Experiments}

We first conduct our experiments on NAS-Bench-201 \cite{dong2021nats} which is a standard benchmark for network topology search. To validate the generalization of our $\ournas$ on large search space, we consider two other size search spaces: AnyNet search space (Sec. \ref{sec:exp_reg}) proposed in \cite{Radosavovic_2020_CVPR} and a new object detection search space (Sec. \ref{sec:exp_yolox}) proposed by us based on YOLOX \cite{ge2021yolox}. Fig. \ref{fig:typeenc}(right) summarizes these three search space. The detailed search space definition can be found in the following subsections.

We adopt SGD with momentum \cite{sutskever2013importance} both for the predictor training and search. During training, we set initial learning rate as 0.01 and decay it with cosine learning rate schedule. We use a weight decay of 0.0001 and a momentum of 0.9. As for search, the learning rate starts from 0.02 and is divided by 10 at 1/3 and 2/3 of the total number of iterations without specific statement.

\subsection*{The Structure of Predictors}
In size search space, we adopt a 3-layer MLP followed by one fully-connected layer as our predictor. As for topology search space, we use a 3-layer GCNs followed by two fully-connected layers. We utilize GeLU \cite{hendrycks2016gaussian} as activation function and  set the dropout as 0.05 in the last fully-connected layers and GCNs layers. Table \ref{tab:struct_predictors} summarizes the structures of predictors. Following \cite{wen2020neural}, each layer in GCNs can be formulated as:
\begin{equation}
    \small
    \label{eq:gcn}
    \MBV_{l+1} = \frac{1}{2}\text{ReLU}(\MBJ \MBV_{l} \MBW_{l}^{1} ) +  \frac{1}{2}\text{ReLU}(\MBJ^{T} \MBV_{l} \MBW_{l}^{2} ),
\end{equation}
where $l$ indicates the index of GCNs layers. $\MBJ \in  \mathbb{R}^{I\times I} $ is the adjacency matrix, $I$ means the number of nodes. The operation matrix is $\MBV_0 \in \mathbb{R}^{I\times D}$, $D$ is the number of operation candidates.  $\MBW_{l}^{1}$ and $\MBW_{l}^{2}$ are trainable weights in GCNs. 

\begin{table}[ht]
    \centering
    \begin{adjustbox}{width=\linewidth}
    \begin{tabular}{c|c|c|c|c|c}
    \toprule
    Predictors & Layers & Embeddings Dimension & Linear Dimension & FC Layers & Activations  \\
    \midrule
    MLP       & 3       & 1000             & 1000   & 1 &  GeLU   \\
    GCNs      & 3       & 144              & 128    & 2 &  GeLU  \\
    \bottomrule
    \end{tabular}
    \end{adjustbox}
    \caption{The structure of predictors in our experiments.} 
    \label{tab:struct_predictors}
\end{table}
\subsection{NAS-Bench-201}
\label{sec:exp_nasbench}
NAS-Bench-201 \cite{dong2019bench} consists of 15625 neural cell candidates with different topologies on CIFAR-10, CIFAR-100 \cite{krizhevsky2009learning} and ImageNet-16-120 \cite{chrabaszcz2017downsampled}. We introduce how we encode the architectures firstly. Following \cite{wen2020neural}, we formulate the structure of cell as a densely connected Directed Acyclic Graph (DAG) and adopt an adjacency matrix and operation matrix to represent the structure. The DAG contains an input node, an output node and up to 6 interior nodes. Each interior node can be one of the following operations: (1) zeroize, (2) skip connection, (3) 1-by-1 convolution, (4) 3-by-3 convolution and (5) 3-by-3 average pooling layer. So the number of nodes in the DAG is $6 + 2 = 8$ and the number of candidate options per node is $5 + 2 = 7$\footnote{Following \cite{wen2020neural}, the input and output are also considered as candidate options.}. Then the topology of cell can be represented by a $8 \times 8$ adjacency matrix. The operation matrix is a $8 \times 7$ matrix which consists of 8 one-hot vectors.
Since there have zeroize and skip connection operations in the candidate operation options, the adjacency matrix can be fixed to a graph with maximum connections in the search space, and only the operation matrix is updated. 

We adopt the same structure of GCN proposed in \cite{wen2020neural} as our predictors. Since this benchmark usually take budget (the number of evaluated models) into consideration, we do not use the resource predictor. As for search, we set the number of iterations to 200 and decay the learning rate by a factor of 2. A softmax operation is applied on each row of the operation matrix to ensure that the sum of each row is 1. After gradient ascent, we select the operator with maximum probability in each row as the chosen operator.

We train main predictors with 30 samples and query performances of the top-40 networks in the model pool, which indicates that \ournas\ trains $30+40=70$ models to get the final performance. Tabel \ref{tab:nb2comparison} shows the results of our method. With the same number of query samples, we achieve better performance than BOHB\cite{falkner2018bohb}, REA\cite{real2019regularized}, RL\cite{zoph2016neural} and a very strong baseline AceNAS\cite{zhang2021acenas} on all the three datasets. On ImageNet-16-120 dataset, our $\ournas$ achieve about $0.6\% - 1.5\%$ higher accuracy than other methods. For fair comparison, we also report the results with 110 query samples following AceNAS. The results are comparable. These results validate the sampling efficiency of our method under low query budget. As shown in the following experiments, this advantage becomes significant when enlarging the search space. 

\begin{table}[t]
    \centering
    \begin{adjustbox}{width=\linewidth}
    \small
    \begin{tabular}{ccllll}
    \toprule
    Method & Querys & CIFAR-10 &CIFAR-100 &  IN-16-120 &  \\
    \midrule
    Optimal$^\ddagger$ & 15625  & 94.37   &  73.51   & 47.31 \\
    \midrule
    Random &  70   &  93.80\std{0.28}  &  71.18\std{1.0}   & 45.38\std{0.91} \\

    BOHB\cite{falkner2018bohb}    & 70 & 93.57\std{0.39} & 71.36\std{0.71} & 45.04\std{0.94} \\
    RL\cite{zoph2016neural}    & 70 & 93.88\std{0.26} & 71.63\std{0.93} & 45.03\std{0.78} \\
    REA\cite{real2019regularized}   & 70 & 94.04\std{0.25}  & 71.78\std{0.90} & 45.52\std{0.65} \\ 
    AceNAS\cite{zhang2021acenas} &   70 & 94.15\std{0.28} & 72.64\std{1.0}  &  45.99\std{0.76} \\
    \ournas &   70  &  94.20\std{0.14} &  73.09\std{0.48}  &  46.72\std{0.35} \\
    \midrule
    AceNAS\cite{zhang2021acenas} &   110 & 94.30\std{0.19} & 73.23\std{0.54}  & 46.47\std{0.38}  \\
    \ournas &   110  &  94.30\std{0.08} &  73.17\std{0.38}  &  46.67\std{0.27} \\

    \bottomrule
\end{tabular}
\end{adjustbox}

\caption{Compared with other NAS methods on NAS-Bench 201. We repeat 15 times to obtain mean and variance. 
$\ddagger$: average results.}
\label{tab:nb2comparison}
\end{table}

\subsection{AnyNet Search Space}
\label{sec:exp_reg}

\begin{table*}[ht]
  \begin{minipage}{\textwidth}
  \scriptsize
  \setlength\extrarowheight{3pt}
  \centering
  \begin{tabular}{l|p{0.15\textwidth}p{0.2\textwidth}p{0.25\textwidth}p{0.18\textwidth}}
  \toprule
   Search Modules & Number ($D_i$) & Widths ($W_i$) & Bottleneck Ratio ($R_i$) & Group ($G_i$)   \\
  \midrule
    Range & $0<D_i\leq16$ & $24^\dagger \leq W_i \leq 1024$  & $\{0.25, 0.5, 1\}$ & $0<G_i\leq32$  \\
  \hline
  \midrule
  RegNetX-600MF & [1,3,5,7] & [48,96,240,528] & [1,1,1,1] & [2,4,10,22]  \\
  \ournas-600MF  & [3,3,6,12] & [64,152,256,1024] & [0.5,0.5,0.5,0.25] & [2,19,32,22]  \\
  \hline
  \end{tabular}
  \caption{AnyNet search space proposed in RegNet\cite{Radosavovic_2020_CVPR}. The below shows the structures of RegNetX-600MF and \ournas-600MF.  $^\dagger$: We increase the lower bound of widths from 0 to 24, and width should be divisible by 8.}
  \label{tab:regnet_ss} 
\end{minipage}
\end{table*}

\begin{table*}[t]
	\centering
	\begin{adjustbox}{width=\linewidth}
		\small
		\begin{tabular}{cccccccccccccccc}
			\toprule
			Training          &       Method   &  Sampled & 200MF & 400MF & 600MF & 800MF & 1.6GF & 3.2GF &  4GF  & 6.4GF & 8.0GF & 12GF  \\
			\midrule 
			\multirow{2}{*}{Proxy} & RegNetX\cite{Radosavovic_2020_CVPR} & $2750$ & 59.3 & 63.2  & 65.1  & 66.1  & 68.9  & 70.0  &  71.4 & 72.6  & 73.4 & 73.7 \\
			& \ournas & $330$  &  59.2 & 63.4  & 65.4  & 66.5  &  69.0 & 70.8  &  71.3 & 72.5  & 73.2 & 73.5  \\ 
						& \ournas$^\dagger$ & $330$  &  58.5 & 63.0  & 65.0  & 66.2  &  69.0 & 70.7  &  71.2 & 72.5  & 73.2 & 73.5  \\ 
			\midrule  
			\multirow{2}{*}{Full} & RegNetX\cite{Radosavovic_2020_CVPR} & $ - $ & 68.9  & 72.7  & 74.1  & 75.2  & 77.0  & 78.3  &  78.6 & 79.2  & 79.3 & 79.7  \\
			& \ournas & $ - $   &67.9 &  72.3  & 73.6  & 74.8  & 76.8  & 77.1  &  78.0 & 79.2  & 79.1 & 79.5 & \\ 
			& \ournas$^\dagger$ & $ - $   &69.0 &  72.5  & 74.2  & 75.0  & 76.8  & 78.1  &  78.4 & 79.2  & 79.1 & 79.5 & \\ 			
			\bottomrule
		\end{tabular}
	\end{adjustbox}
	\caption{Comparison between RegNetX models and our \ournas\ on ImageNet validation set. We report the top-1 accuracy of models with proxy and full training schedule respectively.}
	\label{tab:regnet_comparison}
\end{table*}

We validate our method on ImageNet dataset and compare the searched networks with RegNetX models. RegNetX models are a series networks with different FLOPs obtained from the RegNet search space \cite{Radosavovic_2020_CVPR}. RegNet is obtained by progressively shrinking an unconstrained search space AnyNet by human heuristics. The search space of AnyNet contains 16 variables including the number of blocks, block width, bottleneck ratio and  the number of group in 4 stages. Table \ref{tab:regnet_ss} summarizes the search space of AnyNet, which includes about $(16\cdot128\cdot3\cdot6)^4 \approx 10^{18}$ architectures. The authors of \cite{Radosavovic_2020_CVPR} proposes several prior knowledge to progressively refine the AnyNet search space and sample 500 models every step to validate whether the refinement is worthy. They finally explore a RegNet search space which contains about $10^8$ models after 5 times refinement. Then each of the RegNetX models is obtained by picking the best model from 25 random architectures sampled from the RegNet search space with a specific FLOPs constraints. In this paper, we argue that this strategy still needs intensive human knowledge and trial and error, which is contradict to the spirit of AutoML in fact. Consequently, we adopt our \ournas\ directly on the unconstrained AnyNet search space, and compare the searched architectures to RegNetX models.

We random sample 30 models from the AnyNet search space and train each model for 10 epochs on the ImageNet dataset \cite{deng2009imagenet} following the proxy task setting in \cite{Radosavovic_2020_CVPR}. As \cite{Radosavovic_2020_CVPR} provides RegNetX models with a variety of FLOPs, we adjust the weight of auxiliary predictor $\alpha$ to get the models with different target FLOPs during search. For example, the $\alpha$ is set to 1 for 200MF FLOPs regime while 0.2 for 3.2GF FLOPs. 

For each FLOPs regime, we choose top-30 models from the model pool and pick the best one according to its performance on the proxy task. Then we train the best models with 100 epochs following \cite{Radosavovic_2020_CVPR}. Finally, we sampled $30+30 \times 10=330$ models to get 10 architectures with different FLOPs, which is much less than 2750, the number of models sampled in RegNet \cite{Radosavovic_2020_CVPR}.

As shown in Table \ref{tab:regnet_comparison}, we achieve comparable performances to RegNetX from 200MF to 12GF by directly adopting our method on the AnyNet search space. We found some structures of models we explore are not consistent with the design principles introduced in RegNet. For example,  \ournas-600MF is deeper than RegNetX-600MF (see Table \ref{tab:regnet_ss}), and the bottleneck ratio is not strict to 1. This reveal the fact that human hand-crafted design may still introduce inductive biases which should be avoided. Other architectures of searched model can be found in Appendix.

Another interesting finding is that we find there still exists a non-negligible gap between proxy and full train setting. \ournas\ usually finds significantly better model than RegNetX in the proxy setting, however in full training, the conclusion contradicts. To further validate the gap between proxy and full train setting, we show another series of models PredNAS$^\dagger$ in Table \ref{tab:regnet_comparison}. For each FLOPs regime, We select top-3 models according to their performances on the proxy task, and retrain all of them with 100 epochs. Then the models with best performances are selected as PredNAS$^\dagger$.
The architectures with best performances on full train setting may yield worse results on the proxy task. Consequently, we believe designing a better proxy setting that is more consistent with full training results is a crucial task for efficient NAS, however it is beyond the scope of this work and we leave it for future work.

\subsection{YOLOX Search Space}
\label{sec:exp_yolox}

\begin{table}[ht]
  \centering
  \begin{adjustbox}{width=\linewidth}
  \begin{tabular}{ccccc}
  \toprule
  
  \multicolumn{2}{c}{Search Module}   &   Search Space   & Number \\
  \midrule
  \multirow{1}{*}{Stem} & Width ($S$) & $0 < S \leq 80$ & 10 \\
  \midrule
  \multirow{3}{*}{Backbone} &  Depth ($b$) & $0 < b \leq 12$ & \multirow{3}{*}{$(12\cdot10\cdot3)^4 $ } \\
    & Width ($B^{k}$) & $0<B^{k}\leq B^{k-1} * 2,$ $B^0=80$ &\\
    & Expand ratio  & $\{0.25, 0.5, 1\}$ & \\ 
  \midrule
  \multirow{3}{*}{Neck}  & Depth ($n$) & $0 < n \leq 3$&\\
  & Width ($N^k$) &  $0<N^k\leq 1280$ & \multirow{1}{*}{$(3\cdot10\cdot3)^4 $ } \\
   & Expand ratio & $\{0.25, 0.5, 1\}$  & \\

  \midrule
  \multirow{2}{*}{Shared convs} & Depth ($s$) &  $0 < s \leq 4$& \multirow{2}{*}{$8^{(4+3+2+1)*3} $ } \\ 
   & Width ($W_{s_i}$) & $0<W_{s_i}\leq 512, 0 < s_i \leq s$  &   \\
  \midrule
  \multirow{2}{*}{Cls head} & Depth ($l$) &  $0 < l \leq 4$& \multirow{2}{*}{$8^{(4+3+2+1)*3} $ } \\ 
& Width ($W_{l_i}$) & $0<W_{l_i}\leq 512, 0< l_i \leq l$  &   \\
\midrule
  \multirow{2}{*}{Reg head} & Depth ($r$) &  $0 < r \leq 4$& \multirow{2}{*}{$8^{(4+3+2+1)*3} $ } \\ 
& Width ($W_{r_i}$) & $0<W_{r_i}\leq 512, 0 < r_i \leq r$  &   \\
  \bottomrule
  \end{tabular}
  \end{adjustbox} 
  \caption{The proposed search space of YOLOX. There are four stages in the backbone and neck. We search the depth of each stage. The width and expand ratio of different layers are shared in the $k$-th stage. The width in stem ($S$), backbone ($B^k$) and neck ($N^k$) of each stage will be divided into 10 values. The width in heads will be divided into 8 values. Three decoupled detection heads are responsible for three level of FPN features respectively. In total, the number of architectures in such search space is $10^{100}$. }
  \label{tab:yolox_ss} 
\end{table}

We also adopt our method on the MSCOCO \cite{lin2014microsoft} dataset. The baseline is YOLOX \cite{ge2021yolox}, a highly efficient method proposed very recently. We propose a new search space based on YOLOX. The network architecture in YOLOX consists of three modules: backbone, neck and decoupled detection heads. The backbone and neck are composed of the same type of basic blocks. The searchable dimensions are depth (\textit{i.e.} the number of blocks), widths and expand ratio of blocks. There are three decoupled detection heads in YOLOX, and each of them is responsible for a level of FPN \cite{lin2017feature} feature. The decoupled detection head contains convolution layers (shared convs) followed by two parallel branches for classification (cls head) and regression (reg head), respectively. We search for the depths and widths of convolution layers in each head. The summary of our proposed YOLOX search space is shown in Table \ref{tab:yolox_ss}. This search space includes about $10^{100}$ possible architectures.

\begin{table*}
  \centering
  \footnotesize
  \begin{tabular*}{\textwidth}{l @{\extracolsep{\fill}} lllllllll}
  \toprule
    Model                     & FLOPs (G) & Parameters (M)    & AP (\%) & $\text{AP}_{50}$ & $\text{AP}_{75}$ & $\text{AP}_{S}$ & $\text{AP}_{M}$ & $\text{AP}_{L}$  \\

    \midrule 
    YOLOX-S   &  26.8   & 9.0       & 40.5  & 59.3     & 43.7     & 23.2 & 44.8  & 54.1 \\  
    \ourmodel-S  &  28.4    & 10.6     & \textbf{42.5}  & 61.2     & 46.0     & 23.9 & 47.2  & 55.7 \\  
    \midrule
    YOLOX-M & 73.8   & 25.3   &  46.9   &  65.6     & 51.1      & 29.0    & 52.1    & 62.3   \\
    \ourmodel-M    &  76.0 & 24.1   & \textbf{47.1}  & 66.0 & 51.0 & 28.6 & 52.0 & 62.2 \\ 
    \midrule
    YOLOX-L & 155.6  & 54.2  & 49.7 &  68.0 & 53.9 & 32.2 & 55.0 & 65.1 \\
    \ourmodel-L    & 158.2 & 55.8    &  \textbf{50.5}  &  69.5 & 54.8 & 32.1 & 55.6 & 65.0 \\
    \bottomrule
  \end{tabular*}
  \caption{Comparison between YOLOX and the models searched by our PredNAS in terms of AP ($\%$) on COCO \texttt{val2017}.} 
  \label{tab:yolox_comparison}
\end{table*}

\begin{table}[ht]
  \centering
  \begin{adjustbox}{width=\linewidth}
  \begin{tabular}{ccll}
  \toprule
  \multicolumn{2}{c}{Search Modules}  & YOLOX-S   & \ournas-S \\ 
  \midrule
  Stem & W & 32 & 16 \\
  \midrule
  \multirow{3}{*}{Backbone} &  D  & [1,3,3,1] & [4,3,3,1] \\
                            & W   & [64,128,256,512] & [72,80,320,488] \\
                            & R   & [0.5,0.5,0.5,0.5]   & [0.25,1,0.5,0.25] \\
  \midrule
  \multirow{3}{*}{Neck}     &  D  & [1,1,1,1] & [1,3,1,3] \\
                            & W   & [256,128,256,512]   & [512,256,576,304] \\
                            & R   & [0.5,0.5,0.5,0.5]   & [0.5,0.25,0.5,0.25] \\
  \midrule
  \multirow{2}{*}{Shared convs} &  D  & [1,1,1] & [2,2,3] \\
                              & W   & [[128],[128],[128]]]   & [[224, 40],[224, 176],[256, 136, 176]] \\
  \midrule
  \multirow{2}{*}{Cls head}   &  D  & [2,2,2] & [2,1,3] \\
                              & W   & [[128, 128],[128, 128],[128, 128]]   & [[56, 96],[32], [176, 40, 224]] \\
  \midrule
  \multirow{2}{*}{Reg head}   &  D  & [2,2,2] & [1,2,1] \\
                              & W   & [[128, 128],[128, 128],[128, 128]]   & [[32],[104, 224], [192]] \\                            

\bottomrule
\end{tabular}
\end{adjustbox}
\caption{The structure of YOLOX-S\cite{ge2021yolox} and \ournas-S. W, D and R denote width, depth and ratio, respectively.}
\label{tab:yolox_s} 
\end{table}

For such huge search space, we merely randomly sample 40 models to build training data. We train each sampled model following \cite{ge2021yolox}, but using 6\% train data to alleviate the training cost. 

For each target FLOPs, we select the best architecture from top-30 models in the model pool.

Table \ref{tab:yolox_comparison} shows the results of our searched models. With FLOPs constraints, our models (PredNAS-S/M/L) consistently achieve better performances than YOLOX-S/M/L on the COCO \texttt{val2017} set. Especially, comparing with YOLOX-S, our PredNAS-S achieves 42.5\% AP, improving YOLOX-S by 2.0\%. This result shows the effectiveness of our method on small scale models in object detection task. We further analyze the discrepancy of architectures between YOLOX-S and \ournas-S (see Table \ref{tab:yolox_ss}). \ournas-S has a deeper backbone and computation resources are allocated more on the shared convolutions in head while less on the classification and regression heads.

\section{Discussion}

\begin{figure}[t]\centering
    \begin{minipage}[t]{0.42\linewidth}\resizebox{\columnwidth}{!}{
    \tablestyle{2pt}{1.3}
    \begin{tabular}[b]{ccccc}
        \toprule
        Step & Architecture & Valid-Acc& Pred-Acc \\
        \midrule
        0    & [[4],[2,2],[3,3,3]] & 55.71  & 57.88   \\
        60    & [[4],[2,2],[3,3,3]] & 55.71  & 57.88   \\
        120   & [[4],[2,2],[3,3,1]] & 63.02  & 64.57 \\
        160   & [[2],[2,2],[2,3,1]] & 70.34  & 76.88   \\
        200   & [[2],[2,2],[2,3,1]] & 70.34  & 76.88    \\
        \midrule
        0      & [[5],[1,3],[4,1,2]] & 68.98  & 64.00  \\
        60     & [[5],[1,3],[4,1,2]] & 68.98  & 64.00 \\
        120     & [[5],[1,3],[4,1,2]] & 68.98  & 64.00 \\
        160    & [[2],[1,2],[4,1,2]] & 73.14  & 76.01 \\
        200     &  [[2],[1,2],[4,1,2]] & 73.14  & 76.01 \\
        \bottomrule
    \end{tabular}}\end{minipage}\hspace{.5mm}
    \includegraphics[width=.52\linewidth]{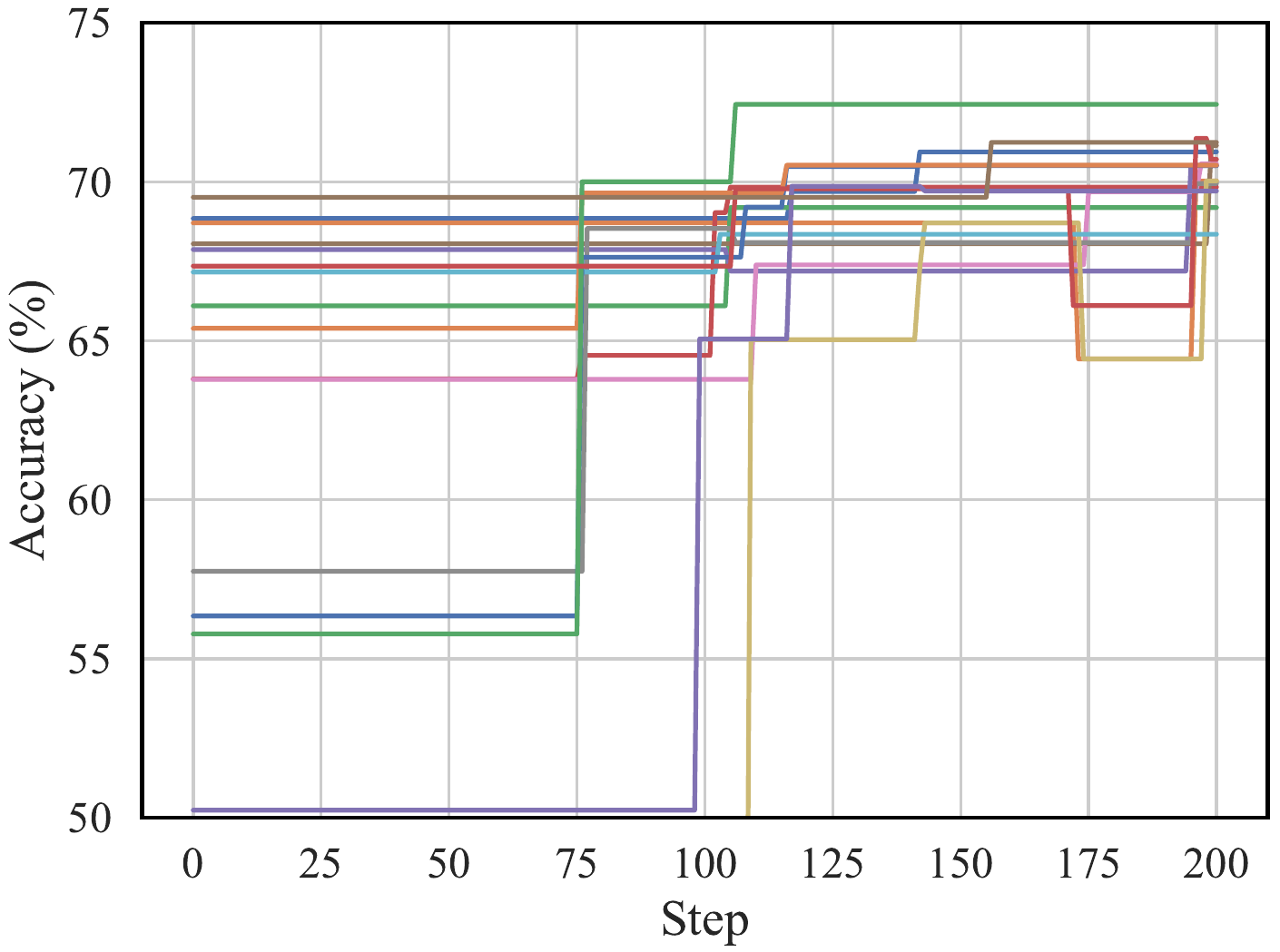}\\
    \caption{\textbf{Intermediate Results on NAS-Bench-201}. \textit{Left:} The intermediate architectures obtained by gradient search on CIFAR100. The index of architecture denote node 1/2/3 and the meaing of value, "1": conv 1x1, "2": conv 3x3, "3": average pool 3x3, "4": skip connection, "5" zeroize  \cite{dong2019bench}. \textit{Right:} The performance of architecture is improving as the step of gradient search increases.}
    \label{fig:vis_gs}\vspace{-3mm}
\end{figure}

\subsection{Effectiveness of Gradient Search}
 \noindent
\textbf{Gradient search can explore better architectures}. In Fig.\ref{fig:vis_gs} (left), we randomly pick some intermediate architectures during the procedure of gradient search on NAS-Bench-201. Given an initial architecture, gradient update can explore novel architecture with better performance. With the increase of prediction score, the accuracy of architecture on the validation set is also improving. 
Fig. \ref{fig:vis_gs} (right) also shows the accuracy curve of models with different initialization. After 200 iterations, most of the architectures achieve 68\%-72\% accuracy at the last step, even with poor initialization. 

\begin{figure}[t]\centering
    \begin{minipage}[t]{0.42\linewidth}\resizebox{\columnwidth}{!}{
    \tablestyle{2pt}{1.2}
    \begin{tabular}[b]{ccc}
        \toprule
        Model & Random & Gradient  \\
        \midrule
        200MF & 56.9\% & 59.2\% \\
        800MF & 65.5\%  & 66.5\% \\
        1.6GF & 67.9\%  & 69.0\% \\
        4GF   & 71.1\%  & 71.3\% \\
        12GF  & 72.3\%  & 73.5\% \\
        \bottomrule
        \vspace{-1mm}
    \end{tabular}}\end{minipage}\hspace{.5mm}
    \includegraphics[width=.52\linewidth]{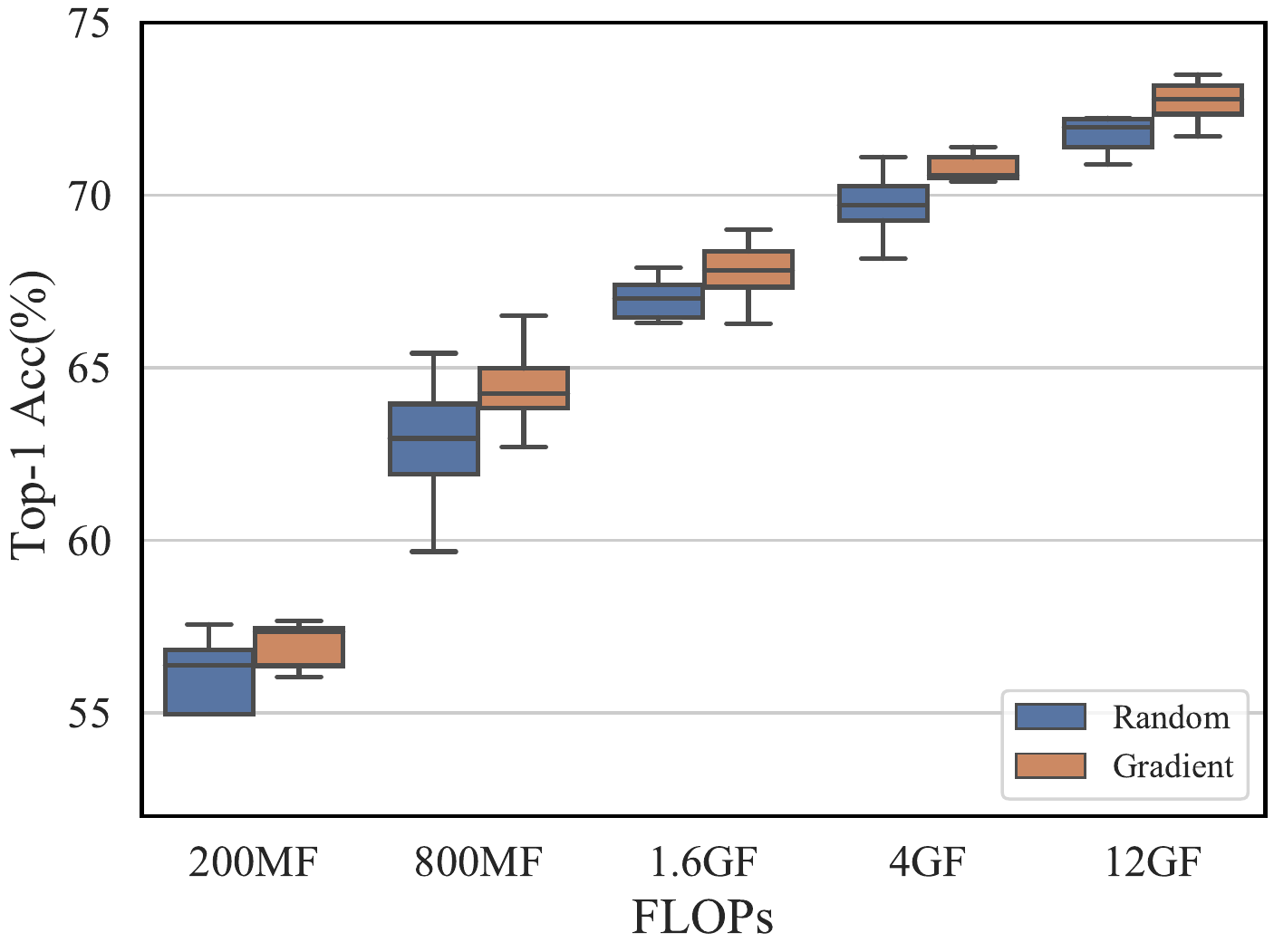}\\
    \caption{Comparison between random search and gradient search. \textit{Left:} The top-1 accuracy of best models on ImageNet. \textit{Right:} The top-1 accuracy of top-15 models in model pool.  All models are trained with our proxy setting.}
    \label{fig:vs_random}\vspace{-2mm}
\end{figure}

\noindent
\textbf{Comparison with random search}. We compare gradient search with random search on the AnyNet search space. If we can traverse all the search space, the result obtained by random search is the same as gradient search due to the identical predictor for ranking models. However, it is unrealistic to traverse all the architectures in a large search space. 
For fair comparison, we randomly sample 100000 models following the number of models explored in gradient search from the AnyNet search space. Then we select top-15 models according to the scores of the main predictor for 5 different FLOPs regimes respectively. For each FLOPs regime, we train the top-15 models on the proxy task and obtain the performance of each model. Fig. \ref{fig:vs_random} (left) shows the best models of different target FLOPs obtained by random search and gradient search. Obviously, gradient search can explore better architecture than random search. Moreover, we show the performances of top-15 models found by random search and gradient search in Fig. \ref{fig:vs_random} (right). Most of the models obtained by gradient search achieve higher accuracy than random search.

\subsection{N + K Ablation Study}
We conduct experiments on NAS-Bench-201 with different N and K. N is the number of training samples for predictor and K is the number of top models selected for evaluation. The top of Table \ref{tab:train_samples} shows the results with different number of training samples. Surprisingly, we found a small number of N is enough for training a predictor which can distinguish the performance of architectures coarsely. With K = 40, training predictor with only 10 samples can achieve better results than several baseline methods shown in Table \ref{tab:nb2comparison}. The bottom of Table \ref{tab:train_samples} shows the results with different K. We found K is more important in our experiments. With the increase of K, the performances are improving stably. This ablation shows that we can train a predictor with few samples and adopt it to do coarse model selection. Then the evaluation of top-K models can be seen as a refinement procedure. 

\begin{table*}[ht]
 \begin{minipage}{\textwidth}
   \centering
   \begin{tabular}{lccccc}
   \hline
   \#Samples N (K = 40) &  10 & 20 & 30 & 40 & 50 \\
   \hline
   CIFAR-100        & 72.85\std{0.70}    &  72.85\std{0.37}  & 73.09\std{0.48} & 72.66\std{0.54} & 72.77\std{0.71} \\
   \hline
   \hline
   \#Top K (N = 30)         &  10 & 20 & 30 & 40 & 50 \\
   \hline
   CIFAR-100  & 71.92\std{0.71} & 72.56\std{0.62} & 72.74\std{0.48} & 73.09\std{0.48} & 73.17\std{0.27} \\ 
   \hline
   \end{tabular}
   \caption{The results of \ournas\ with different N and K on NAS-Bench-201. We repeat 15 times to obtain mean and variance.}
   \label{tab:train_samples}
\end{minipage}
\end{table*}

\section{Conclusion}
In this paper, we have proposed a universal framework named $\ournas$ for neural architecture search. In our framework, we adopted neural predictor to estimate the performance of architectures and used a simple gradient search strategy to do architecture search. We validate our method on NAS-Bench-201, ImageNet and MSCOCO. With less than 100 training samples, our $\ournas$ could achieve comparable or even better performances than existing state-of-the-art methods. To reduce the training cost, we adopted proxy task to get the performance of architectures. However, we found the consistency of model performances between proxy task and the final task would limit the effectiveness of our method. It is interesting to investigate how to design the proxy task by the technique of NAS, which we leave it as our future work. 

{\small
\bibliographystyle{ieee_fullname}
\bibliography{egbib}
}

\newpage
\clearpage
\section*{Supplementary}
\label{sec:supp}

\subsection*{Searched Models}
\label{sec:apdx_search}
\textbf{RegNet.} Table \ref{tab:regnet_search_all} shows the searched models of PredNAS and PredNAS$^\dagger$ in RegNet search space. Following to RegNet, we increase the upper bound of search space to get large scale models during search. Therefore, some dimensions in the searched models may outrange the AnyNet search space described in the main paper. For example, the width of \ournas-12GF in the last stage is 1368, larger than 1024. Interestingly, most of the searched models have increasing widths as the RegNet suggested. 

\noindent
\textbf{YOLOX.} We adjust the lower bound to obtain models with small FLOPs during search. Table \ref{tab:yolox_search_all} describes the models we found in the YOLOX search space.
\ournas-S/M/L are deeper than YOLOX-S/M/L and the proportion of share convs is increasing. 

\begin{table*}
    \centering
    \footnotesize
    \begin{adjustbox}{width=\linewidth}
    \begin{tabular}{clllllll}
    \toprule
    Search Module  &  Depth &  Width & Ratio  & Groups$^*$  & Group Width & Flops(B)  & Acc(\%)  \\ 
    \midrule
    RegNetX-200MF  & [1,1,4,7] & [24,56,152,368] & [1,1,1,1] & [3,7,9,46] &  8  & 0.2 & 68.9 \\
    \ournas-200MF  & [5,2,16,1] & [24,32,96,1600] & [1,1,1,0.25] & [3,1,12,16] & [8,32,8,25] & 0.22 & 67.9 \\
    \ournas$^\dagger$-200MF  & [2,11,13,14] & [24, 88, 56, 616] & [0.5,0.25,1,0.25] & [6,11,8,22] & [2,2,7,7] & 0.22 & 69.0 \\
    \midrule
    RegNetX-400MF  & [1,2,7,12] & [32,64,160,384] & [1,1,1,1] & [2,4,10,24]& 16  & 0.4 & 72.7 \\
    \ournas-400MF & [4,2,10,8] & [64,232,64,760] & [0.5,0.25,1,0.5] & [8,29,32,38] &[4,2,2,10]   & 0.41 & 72.3 \\
    \ournas$^\dagger$-400MF & [3,3,15,8] & [64,64,64,792] & [0.5,0.5,1,0.5] & [4,8,8,33] & [8,4,8,12]  & 0.4 & 72.5 \\
    \midrule
    RegNetX-600MF  & [1,3,5,7] & [48,96,240,528] & [1,1,1,1] & [2,4,10,22] & 24 & 0.6 & 74.1\\
    \ournas-600MF  & [3,3,6,12] & [64,152,256,1024] & [0.5,0.5,0.5,0.25] & [2,19,32,32]& [16,4,4,8]& 0.61 & 73.6\\
    \ournas$^\dagger$-600MF & [13,2,11,8] & [32,128,208,880] & [0.5,1,0.5,0.5] & [16,8,26,22]& [1,16,4,20]& 0.61 & 74.2\\
    \midrule
    RegNetX-800MF  & [1,3,7,5] & [64,128,288,672] & [1,1,1,1] & [4,8,18,42] & 16 &0.8 & 75.2 \\
    \ournas-800MF  & [6,3,13,2] & [96,168,168,1048] & [0.25,0.5,1,1] & [8,4,28,8]&[3,21,6,131] & 0.77 & 74.8 \\
    \ournas$^\dagger$-800MF  & [4,9,13,11] & [64,64,168,888] & [0.5,1,0.5,0.5] & [16,4,12,37]&[2,16,7,12] & 0.76 & 75.0 \\
    \midrule
    RegNetX-1.6GF  & [2,4,10,2] & [72,168,408,912] & [1,1,1,1] & [3,7,17,38]&24  & 1.6 & 77.0 \\
    \ournas-1.6GF  & [4,1,8,11] & [128,240,472,968] &[0.5,0.5,0.5,0.5] & [8,10,4,11]&[8,12,59,44] & 1.64 & 76.8 \\
    \ournas$^\dagger$-1.6GF  & [4,1,8,11] & [128,240,472,968] &[0.5,0.5,0.5,0.5] & [8,10,4,11]&[8,12,59,44] & 1.64 & 76.8\\
    \midrule
    RegNetX-3.2GF  & [2,6,15,2] & [96,192,432,1008] & [1,1,1,1] & [2,4,9,21] &32  & 3.2 & 78.3\\
    \ournas-3.2GF  & [5,1,14,13] & [128,128,496,1144] & [0.5,0.5,0.5,1] & [32,16,31,26] &[2,4,8,44] & 3.1 & 78.3\\
    \ournas$^\dagger$-3.2GF  & [12,1,16,6] & [128,552,528,984] & [0.5,0.5,0.5,1] & [8,46,33,12] &[8,6,8,82]  & 3.0 & 78.1\\ 
    \midrule
    RegNetX-4.0GF  & [2,5,14,2] & [80,240,560,1360] & [1,1,1,1] & [2,6,14,34] & 40 & 4.0 & 78.6\\
    \ournas-4.0GF  & [13,5,14,1] & [128,128,992,1024] & [0.5,0.5,0.5,1] & [32,16,31,4] &[2,4,16,256] & 4.1 & 78.0\\
    \ournas$^\dagger$-4.0GF  & [10,2,19,10] &[88,152,704,1368]& [1,0.5,0.5,0.5]&[8,19,32,12] & [11,4,11,57]& 4.0 & 78.4\\
    \midrule 
    RegNetX-6.4GF  & [2,4,10,1] & [168,392,784,1624] & [1,1,1,1] & [3,7,14,29] &56 & 6.5 & 79.2 \\
    \ournas-6.4GF & [12,2,17,1] & [184,296,1240,1316] & [0.5,0.5,0.25,1] & [4,37,10,1] &[23,4,31,1316] & 6.4 & 79.2\\
    \ournas$^\dagger$-6.4GF & [12,2,17,1] & [184,296,1240,1316] & [0.5,0.5,0.25,1] & [4,37,10,1] &[23,4,31,1316] & 6.4 & 79.2\\
    \midrule  
    RegNetX-8.0GF  & [2,5,15,1] & [80,240,720,1920] & [1,1,1,1] & [1,2,6,16] & 80 & 8.0 & 79.3 \\
    \ournas-8.0GF  & [8,2,15,1] & [304,512,1184,1408]&[0.25,0.5,0.5,1] & [38,16,37,1] & [2,16,16,1408] & 7.9 & 79.1\\
    \ournas$^\dagger$-8.0GF  & [8,2,15,1] & [304,512,1184,1408]&[0.25,0.5,0.5,1] &  [38,16,37,1] & [2,16,16,1408] & 7.9 & 79.1\\
    \midrule  	
    RegNetX-12GF  & [2,5,11,1] & [224,448,896,2240] & [1,1,1,1] & [2,4,8,20] &  112 & 12.1 & 79.7\\
    \ournas-12GF  & [13,2,16,2] & [136,720,1184,1368] & [1,0.5,0.5,1] & [8,2,8,19] &[17,160,74,72] & 11.8 &79.5 \\
    \ournas$^\dagger$-12GF  & [13,2,16,2] &[136,720,1184,1368] & [1,0.5,0.5,1] & [8,2,8,19] &[17,160,74,72] & 11.8 &79.5 \\
    \bottomrule
    \end{tabular}
    \end{adjustbox}
    \caption{Searched Models in RegNet Search Space. $^*$: In gradient search, we will convert the group to the nearst integer of itself making it can divided by the input channel. Group Width is equal to the input channel divided by group.}
    \label{tab:regnet_search_all} 
\end{table*}

\label{sec:apdx_yolox}

\begin{sidewaystable}
    \centering
    \footnotesize
    \begin{adjustbox}{width=\linewidth}
    \begin{tabular}{clll|lll|ll|ll|ll}
    \toprule
    Search Modules &  \multicolumn{3}{c}{Backbone}  & \multicolumn{3}{c}{Neck}  &  \multicolumn{2}{c}{Share Conv} &  \multicolumn{2}{c}{Cls Head} &  \multicolumn{2}{c}{Reg Head} \\
    \midrule
    & depth & width & ratio & depth  & width & ratio & depth & width & depth & width & depth & width   \\
    \midrule
    YOLOX-S & $[1,3,3,1]$ & $[32,64,128,256,512]$ & $[0.5,0.5,0.5,0.5]$ & $[1,1,1,1]$ & $[256,128,256,512]$ & $[0.5,0.5,0.5,0.5]$ & $[1,1,1]$ & $[[128],[128],[128]]$ & $[2,2,2]$ & $[[128, 128],[128, 128],[128, 128]]$ & $[2,2,2]$ & $[[128, 128],[128, 128],[128, 128]]$ \\
    \ournas-S & $[4,3,3,1]$ & $[16,72,80,320,488]$ & $[0.25,1,0.5,0.25]$ & $[1,3,1,3]$ & $[512,256,576,304]$ & $[0.5,0.25,0.5,0.25]$ & $[2,2,3]$ & $[[224, 40],[224, 176],[256, 136, 176]]$ & $[2,1,3]$ & $[[56, 96],[32], [176, 40, 224]]$ & $[1,2,1]$ & $[[32],[104, 224], [192]]$ \\ 
    \ournas-S$^T$ & $[1,1,1,6]$ & $[8,48,96,256,336]$ & $[0.25,0.5,0.25,0.25]$ & $[1,2,3,3]$ & $[688,288,788,472]$ & $[0.25,0.25,0.25,1]$ & $[1,1,1]$ & $[[32],[128],[96]]$ & $[1,1,3]$ & $[[32],[120], [32,248,224]]$ & $[2,2,4]$ & $[[64,32],[232, 80], [32,112,144,144]]$ \\ 
    \midrule
    YOLOX-M & $[2,6,6,2]$ & $[48,96,192,384,768]$ & $[0.5,0.5,0.5,0.5]$ & $[2,2,2,2]$ & $[384,192,384,768]$ & $[0.5,0.5,0.5,0.5]$ & $[1,1,1]$ & $[[192],[192],[192]]$ & $[2,2,2]$ & $[[192,192],[192,192],[192, 192]]$ & $[2,2,2]$ & $[[192,192],[192,192],[192,192]]$ \\
    \ournas-M & $[3,8,6,2]$ & $[48,96,192,384,768]$ & $[0.25,0.25,0.5,0.25]$ & $[2,2,3,2]$ & $[384,192,640, 768]$ & $[0.5,0.25,0.25,0.25]$ & $[3,1,3]$ & $[[192,248,192],[192],[192,192,192]]$ & $[2,2,2]$ & $[[192,192],[192,240],[192,192]]$ & $[2,2,3]$ & $[[192,192],[192,192],[192,232,208]]$  \\
    \ournas-M$^T$ & $[1,1,10,2]$ & $[64,96,216,512,664]$ & $[0.5,0.5,0.25,0.5]$ & $[1,3,3,1]$ & $[752,128,1024,976]$ & $[0.25,0.5,0.5,1]$ & $[1,1,3]$ & $[[32],[32],[488,312,160]]$ & $[1,1,4]$ & $[[32,],[48],[104,408,456,480]]$ & $[1,4,4]$ & $[[168],[32,320,208,344],[328,232,160,376]]$  \\
    \midrule
    YOLOX-L & $[3,9,9,3]$ & $[64,128,256,512,1024]$ & $[0.5,0.5,0.5,0.5]$ & $[3,3,3,3]$ & $[512,256,512,1024]$ & $[0.5,0.5,0.5,0.5]$ & $[1,1,1]$ & $[[256],[256],[256]]$ & $[2,2,2]$ & $[[256, 256],[256, 256],[256, 256]]$ &  $[2,2,2]$ & $[[256, 256],[256, 256],[256, 256]]$  \\
    \ournas-L & $[12,12,11,10]$ & $[48,120,240,384,960]$ & $[0.5,0.5,0.5,0.5]$ & $[3,2,2,3]$ & $[384,568,640,768]$ & $[0.25,0.25,0.5,0.25]$ & $[3,1,3]$ & $[[192,208,192],[256],[192,256,192]]$ & $[3,2,3]$ & $[[192,192,192],[192,256],[256,256,256]]$ &  $[2,3,2]$ & $[[192,192],[256,256,256],[256,192]]$ \\

    \bottomrule
    \end{tabular}
    \end{adjustbox}
    \caption{Searched models in YOLOX search space.}
    \label{tab:yolox_search_all} 
\end{sidewaystable}

\end{document}